\documentclass[conference, a4paper]{IEEEtran}
\IEEEoverridecommandlockouts
\makeatletter
\newcommand{\newlineauthors}{%
  \end{@IEEEauthorhalign}\hfill\mbox{}\par
  \mbox{}\hfill\begin{@IEEEauthorhalign}
}
\usepackage{cite}
\usepackage{amsmath,amssymb,amsfonts}
\usepackage{algorithmic}
\usepackage{graphicx}
\usepackage{textcomp}
\usepackage{xcolor}
\usepackage{multirow,multicol, makecell, booktabs}
\usepackage[hyphens]{url}
\usepackage{breakurl}
\def\BibTeX{{\rm B\kern-.05em{\sc i\kern-.025em b}\kern-.08em
    T\kern-.1667em\lower.7ex\hbox{E}\kern-.125emX}}
\begin{document}

\makeatletter
\def\ps@IEEEtitlepagestyle{%
  \def\@oddfoot{\mycopyrightnotice}%
  \def\@evenfoot{}%
}
\def\mycopyrightnotice{%
  {\footnotesize 978-8-8872-3749-8 \copyright 2020 AEIT \hfill} 
}
\makeatother

\title{Visual Saliency Detection in Advanced Driver Assistance Systems \\

}

\author{\IEEEauthorblockN{Francesco Rundo}
\IEEEauthorblockA{\normalsize{ADG, Central R\&D}\\
\textit{STMicroelectronics}\\
Catania, Italy \\
francesco.rundo@st.com}
\and
\IEEEauthorblockN{Michael Sebastian Rundo}
\IEEEauthorblockA{\textit{Perceive Lab}\\
\textit{University of Catania} \\
Catania, Italy \\
michael.rundo3764@gmail.com }
\and
\IEEEauthorblockN{Concetto Spampinato}
\IEEEauthorblockA{\textit{Perceive Lab}\\
\textit{University of Catania} \\
Catania, Italy \\
cspampinato@dieii.unict.it }

}

\maketitle

\begin{abstract}
Visual Saliency refers to the innate human mechanism of focusing on and extracting important features from
the observed environment. Recently, there has been a notable
surge of interest in the field of automotive research regarding
the estimation of visual saliency. While operating a vehicle,
drivers naturally direct their attention towards specific objects,
employing brain-driven saliency mechanisms that prioritize certain elements over others. In this investigation, we present an
intelligent system that combines a drowsiness detection system for
drivers with a scene comprehension pipeline based on saliency. To
achieve this, we have implemented a specialized 3D deep network
for semantic segmentation, which has been pretrained and
tailored for processing the frames captured by an automotive-grade external camera. The proposed pipeline was hosted on
an embedded platform utilizing the STA1295 core, featuring
ARM A7 dual-cores, and embeds an hardware accelerator.
Additionally, we employ an innovative biosensor mounted on the
steering wheel to monitor the driver’s drowsiness, gathering the
PhotoPlethysmoGraphy (PPG) signal. A dedicated 1D temporal
deep convolutional network has been devised to classify the
collected PPG time-series, enabling us to assess the driver’s level
of attentiveness. Ultimately, we compare the determined attention
level of the driver with the corresponding saliency-based scene
classification to evaluate the overall safety level. The efficacy
of the proposed pipeline has been validated through extensive
experimental results.
\end{abstract}


\section{Introduction}
Drowsiness denotes a physiological state characterized by reduced consciousness and challenges in maintaining wakefulness, posing a significant risk for road accidents. Consequently, numerous studies have explored methods to monitor the driver's attentiveness. These investigations have identified a correlation between human attention, often assessed through drowsiness monitoring, and the Heart Rate Variability (HRV) index. HRV quantifies heart activity across consecutive beats, yielding valuable insights into drowsiness levels [1]. Specifically, HRV serves as a non-invasive means of assessing the activity of the Autonomous Nervous System (ANS) [1]. The scientific literature extensively recognizes the link between an individual's attention level and ANS activity [1]-[3]. Hence, by analyzing HRV, we can indirectly gauge the subject's level of attentiveness. However, it is crucial to consider the current driving situation when evaluating the driver's attention. For instance, driving in low-traffic and low-speed conditions imposes less cognitive demand compared to scenarios involving risky maneuvers such as overtaking or lane changes. This study introduces an algorithm that integrates saliency analysis, specifically for comprehending driving scenes, with physiological monitoring of the car driver to assess their attention level. Saliency analysis has found extensive utility across various domains, including automotive applications, enabling the identification of crucial elements within videos or images  \cite{cai2017saliency, winterlich2013saliency}.However, the algorithms for analyzing driving scenes and monitoring drowsiness impose substantial computational demands. Through the proposed saliency-based approach, it becomes possible to characterize the driving scene by focusing solely on salient elements such as vehicles and pedestrians. This not only reduces the computational load but also enhances algorithmic performance. The subsequent sections of this paper are structured as follows: Section II provides a review of previous works in the field, exploring the utilization of physiological signal analysis for automated drowsiness detection. Section III presents a detailed description of the constituent blocks comprising our pipelines. Section IV outlines our experimental setup, including information regarding the dataset and hardware devices utilized in implementing our pipeline. Finally, Section V concludes the paper, summarizing the findings and outlining potential directions for future research.

\begin{figure*}
\centerline{\includegraphics[ width=0.7\textwidth]{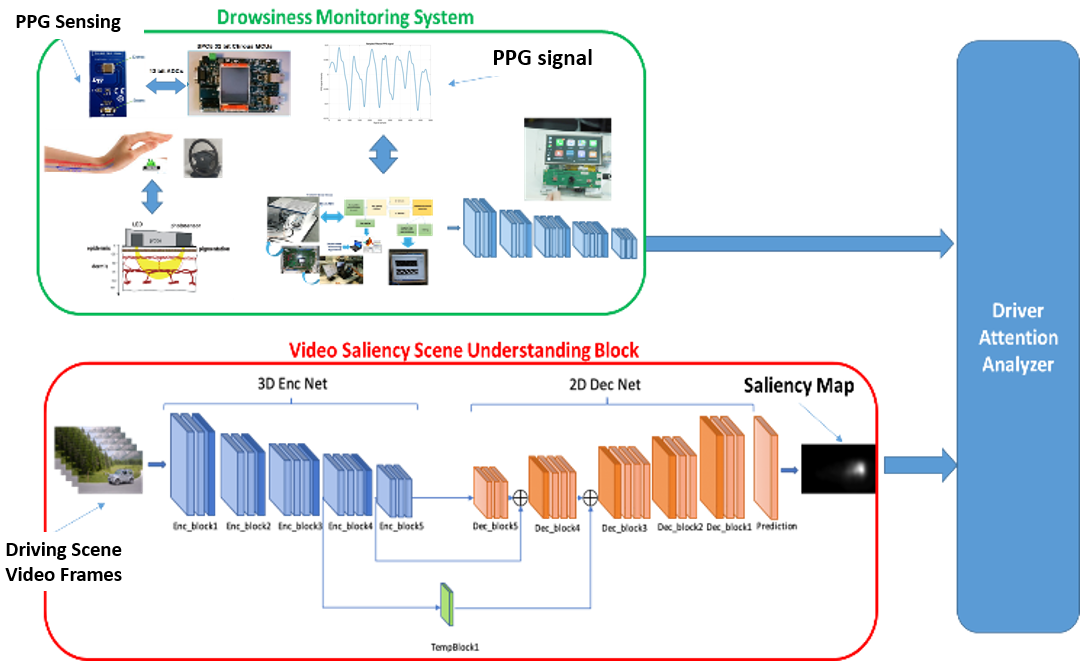}}
\caption{The overall scheme of the proposed pipeline.}
\label{fig1}
\end{figure*}

\section{Related Works}
Numerous research endeavors have focused on developing effective methods for simultaneously monitoring the driver's attention level and analyzing driving scenarios. In a notable study \cite{vicente2011detection}, the researchers devised a pipeline for detecting driver drowsiness by capturing the ElectroCardioGraphy (ECG) signal and analyzing alterations in Heart Rate Variability (HRV). Through ECG signal stabilization and employing classical linear discriminant analysis, the pipeline successfully discriminates between drowsy and wakeful states. However, a key challenge in drowsiness detection lies in ECG sampling. The minimal setup for acquiring a subject's ECG signal necessitates the use of at least three electrodes in contact with the skin, following the configuration known as Einthoven's Triangle ~\cite{abi2019einthoven}. Consequently, the driver must place their hands on electrodes embedded in the steering wheel to capture the ECG signal, with a third electrode typically positioned in the driver's seat. Unfortunately, this placement increases susceptibility to noise, thereby affecting the reliability of the HRV signal derived from the collected ECG signal \cite{abi2019einthoven}. To address these limitations, several researchers have turned to analyzing the PhotoPlethysmoGraphic (PPG) signal, which offers a more favorable alternative. The PPG signal requires only a single contact point for sampling, making it more practical compared to ECG \cite{rundo2018advanced}.  In another study~\cite{rundo2018advanced}, he authors explored parasympathetic nervous activity by extracting HRV features from the PPG signal, aiming to classify the driver's fatigue level. The obtained results demonstrated the robustness of their proposed approach. Moreover, in a separate study \cite{rundo2018nonlinear},  the authors put forth a reliable indicator derived from the PPG signal. 
In previous studies [1][2], researchers have reported encouraging outcomes by employing algorithms that leverage Pulse Rate Variability (PRV) data processing. These investigations have demonstrated the utility of PRV as a valuable metric for analyzing the Autonomous Nervous System (ANS) and evaluating subject drowsiness. Another noteworthy study by Ryu et al. [3] employed red organic light-emitting diodes (OLEDs) and organic photodiodes (OPDs) to capture the photoplethysmographic (PPG) signal. The evaluation of this flexible PPG sensor showcased its effectiveness in detecting drowsiness, surpassing the performance of standard PPG probes in terms of accuracy. In recent years, Deep Learning approaches have garnered significant attention for drowsiness estimation, employing bio-signals and image analysis. For instance, in studies by Hong et al. [4] and Alshaqaqi et al. [5], methods based on image processing were implemented to track the driver's face, gaze, and emotions to assess their attentiveness and drowsiness. However, it is important to note that image processing performance can be influenced by various factors in the in-vehicle environment, such as lighting conditions and occlusions, which can impact the accuracy of the analysis. As highlighted, the automotive industry's continual technological advancements have led to the emergence of promising Drowsiness Detection systems, as exemplified by the pipeline proposed in [6]. Furthermore, a range of Deep Learning algorithms has been devised by researchers to assess the level of fatigue by employing "data fusion" techniques that combine visual and physiological information from the car driver. Altun and Celenk [1] introduced a vision-based driver assistance system focused on scene awareness, leveraging visual saliency analysis. Their results demonstrated the system's effectiveness in enhancing the driver's situational awareness (SA) by facilitating adaptive road surface classification. In another study [2], Deng et al. investigated driver drowsiness by conducting eye-tracking experiments during driving. They recruited a group of 40 subjects, including both non-drivers and experienced individuals. Their findings indicated that drivers exhibit heightened attention towards the vanishing points on the road. This observation suggests that the road's vanishing point could serve as a valuable cue for estimating the road area in a traffic saliency detection mode. To incorporate drowsiness monitoring, the authors proposed an integrated solution that extracts relevant information regarding the driving scenario [2].


\section{Methods and Materials}
This section delves into the finer aspects of the proposed pipeline, illustrated in Figure 1.
\subsection{The Attention Monitoring System}
Our study involved the implementation of a system designed to capture the car driver's Photoplethysmographic (PPG) signal, which enables us to assess and monitor their attention level. As mentioned earlier, the PPG signal offers a non-invasive means of analyzing the heart's pulse rate. Notably, it provides valuable insights into cardiac and vascular conditions, as well as respiratory rate \cite{rundo2018advanced}. The PPG waveform comprises two key components: an 'AC' physiological signal that correlates with cardiac-synchronous changes in blood volume, and a 'DC' component that reflects minor variations resulting from respiration and thermoregulation processes \cite{rundo2018advanced}.
As the heart propels blood towards the periphery, it generates pressure, leading to the expansion of arteries and arterioles in the subcutaneous tissue. To capture the Photoplethysmographic (PPG) signal, we employed a device comprising a light-emitting component and a detector that made contact with the subject's skin. By illuminating the skin and gauging the amount of light scattered back, we could effectively detect the fluctuations in volume that occur with each heartbeat. These variations are manifested as peaks within the PPG waveform [1]. The heart's pressure pulses, as it pumps blood to the periphery, induce specific distension within the subcutaneous arteries and arterioles. Consequently, our approach involved employing a device with a light-emitting component and a detector in direct contact with the subject's skin, facilitating the sampling of the PPG signal. By illuminating the skin and accurately measuring the amount of light scattered back, we were able to identify the changes in volume associated with each heart pressure pulse. These changes are graphically represented as peaks within the PPG waveform. ~\cite{rundo2018advanced}. 
We used the PPG sampling device composed of the Silicon Photomultiplier sensor \cite{vinciguerra2018ppg, mazzillo2018characterization, rundo2019innovative, rundo2019ad}. The proposed PPG probes show an array detector device, called Silicon Photomultipliers (SiPMs)~\cite{mazzillo2018characterization}, with a total area of $4.0\times4.5$ $mm^{2}$ and $4871$ square microcells with $60$ µm pitch. The devices have a geometrical fill factor of $67.4\%$ and are packaged in a surface mount housing (SMD) with about $5.1\times5.1$ $mm^{2}$ total area \cite{fujiwara2018heart}. We propose a Pixelteq dichroic bandpass filter with a pass-band centered at about $540$ nm with a Full Width at Half Maximum (FWHM) of $70$ nm and an optical transmission higher than $90-95\%$ in the pass-band range was glued on the SMD package by using a Loctite 352TM adhesive. The SiPM has a maximum detection efficiency of about $30 \%$ at 565 nm and a PDE of about $27.5\%$ at $540$ nm (central wavelength in the filter pass-band). As described, the PPG detector is composed of a light emitter in combination with a detector based on SiPM technology. The OSRAM LT M673 LEDs were used by SMD package and InGan technology \cite{fujiwara2018heart}. The used LEDs devices are characterized by an area of $2.3\times1.5$ $mm^{2}$ with a 120° angle view, a spectral bandwidth of $33$ nm and a lower power emission (mW) in the standard range. In order to enhance the efficiency and utility of the PPG probe, we developed a printed circuit board (PCB) with a user-friendly interface utilizing National Instruments (NI) instrumentation. The PCB incorporates various components such as a 4V portable battery, power management circuits, a conditioning circuit for output SiPMs (Silicon Photomultipliers) signals, and multiple USB connectors to accommodate PPG probes. Additionally, the board features SMA (SubMiniature version A) output connectors for seamless connectivity and integration with other devices and systems.
\begin{table}
\caption{HYPER LOW-PASS FILTERING SETUP (IN HZ)}
\begin{center}
\resizebox{\columnwidth}{!}{%
\begin{tabular}{|c|c|c|c|c|c|c|c|c|c|c|c|}
\hline
\rule{0pt}{3ex}
\large\textbf{F} & \large\textit{f1} & \large\textit{f2} & \large\textit{f3} & \large\textit{f4}& \large\textit{f5} & \large\textit{f6}& \large\textit{f7} & \large\textit{f8}& \large\textit{f9} & \large\textit{f10}& \large\textit{f11} \\
\hline
\rule{0pt}{3ex}
\large HP & \large0.5 & \large/ & \large/ & \large/ & \large/ & \large/ & \large/ & \large/ & \large/ & \large/ & \large/ \\
\hline
\rule{0pt}{3ex}
\large LP & \large 0 & \large 1.4 & \large 2.9 & \large 2.5 & \large 3.8 & \large 3.9 & \large 4 & \large 4.5 & \large 5 & \large 5.3 & \large 6.9 \\
\hline
\end{tabular}}
\label{tab1}
\end{center}
\end{table}

\begin{table}
\caption{HYPER HIGH-PASS FILTERING SETUP (IN HZ)}
\begin{center}
\resizebox{\columnwidth}{!}{%
\begin{tabular}{|c|c|c|c|c|c|c|c|c|c|c|c|}
\hline
\rule{0pt}{3ex}
\large\textbf{F} & \large\textit{f1} & \large\textit{f2} & \large\textit{f3} & \large\textit{f4}& \large\textit{f5} & \large\textit{f6}& \large\textit{f7} & \large\textit{f8}& \large\textit{f9} & \large\textit{f10}& \large\textit{f11} \\
\hline
\rule{0pt}{3ex}
\large HP & \large0.5 & \large1.2 & \large2.6 & \large2.7 & \large3.3 & \large3.5 & \large4 & \large4.4 & \large5 & \large5.7 & \large6.4  \\
\hline
\rule{0pt}{3ex}
\large LP & \large7 & \large/ & \large/ & \large/ & \large/ & \large/ & \large/ & \large/ & \large/ & \large/ & \large/ \\
\hline
\end{tabular}}
\label{tab1}
\end{center}
\end{table}

In our previous works [1][2][3][4], we provided further insights into the hardware setup utilized for acquiring the Photoplethysmographic (PPG) signal. The PPG Sensor Probe consists of the SiPM sensor along with the aforementioned LEDs. Effective management of the SiPM device's power consumption is facilitated by a dedicated Power management circuit [1][2][3].
As previously mentioned, multiple PPG probes were strategically positioned on the steering wheel of the vehicle to capture the PPG signal. It suffices for the driver to place only one hand over the PPG sensor to collect the physiological signal accurately. Once the raw PPG signal is acquired, the National Instruments (NI) device processes the collected data using an internal $24$ bit analog-to-digital converter (ADC). Additionally, the NI device is equipped with a Windows-based operating system, incorporating a LabView software framework [4]. 
To preprocess the acquired PPG raw data, we devised a LabView algorithm that incorporates several key steps. Firstly, we applied ad-hoc Finite Impulse Response (FIR) filtering techniques, encompassing both low-pass and high-pass bands, to refine the signal quality. Additionally, we performed further processing by calculating the first and second derivatives of the PPG signal, enabling us to evaluate the minimum and maximum values for each waveform accurately. Lastly, we visualized the processed PPG signal on the monitor connected to the National Instruments (NI) device, providing real-time feedback and visualization of the results. 
Further details regarding the National Instruments (NI) device and the LabView software framework can be found in our previous works [1][2]. However, it is important to note that the final implementation of the proposed pipeline will be transitioned to the STA1295 embedded platform, replacing the NI device utilized solely for development purposes [3][4]. This transition ensures a more streamlined and optimized solution.
For the purpose of this study, we specifically developed an efficient pipeline for processing the raw Photoplethysmographic (PPG) data using the MATLAB framework, as described in the validation session of our research. This approach allowed us to effectively analyze and validate the performance of our pipeline. 
Table I and II present comprehensive information about the hyper-filtering configuration employed for processing the recorded raw Photoplethysmographic (PPG) signal [1][2][3]. The hyper-filtering technique, introduced by the authors, draws inspiration from hyper-spectral imaging and applies it to signal processing [3]. By employing hyper-filtering, the source PPG signal undergoes filtering and processing with various frequency setups (as outlined in the tables) to extract discriminative patterns associated with the drowsiness level of the subject under analysis. Further details regarding this approach can be found in [3]. Once the hyper-filtered PPG signal patterns have been collected [2], they are classified using a specialized 1D Temporal Dilated Convolutional Neural Network (1D-CNN). Our proposed Temporal Convolutional Network architecture incorporates a Dilated Causal Convolution layer that operates on the temporal sequence of input data. The "causal" aspect ensures that the activations at time $t$ are solely influenced by inputs from time $t - 1$. The network architecture comprises multiple residual blocks, each consisting of two dilated causal convolution layers with a consistent dilation factor, followed by normalization, ReLU activation, and spatial dropout layers. Specifically, our implemented 1D-CNN consists of 12 blocks, culminating in a downstream softmax layer. Each block consists of a dilated convolution layer with $3\times3$ kernel filters, a spatial dropout layer, another dilated convolution layer, a ReLU layer, and a final spatial dropout layer. The initial dilation size is set to 2, incrementing it for each subsequent block [3]. The objective of this study is to predict the drowsiness level of the car driver by processing the hyper-filtered signal patterns derived from the acquired PPG signal using the designed 1D-CNN architecture. The network's output is a scalar value ranging from 0 to 1, indicating the driver's attention level and distinguishing between drowsy (0) and wakeful (1) states. Our findings strongly support the feasibility of accurately estimating driver drowsiness. To ensure effective support during driving, we have successfully ported the proposed 1D Deep CNN backbone to an automotive-grade STA1295 Accordo5 embedded platform [4]


\begin{figure}
\centerline{\includegraphics[ width=0.8\columnwidth]{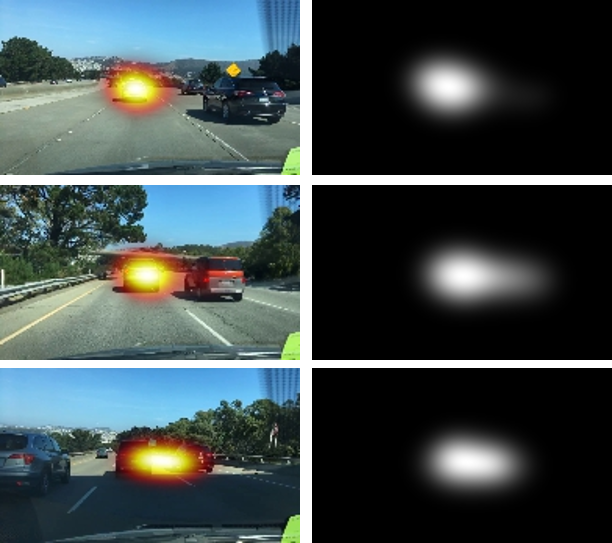}}
\caption{Saliency analysis of the video representing the driving scene.}
\label{fig3}
\end{figure}

\subsection{The Video Saliency Scene Understanding Block}
This section provides detailed insights into the architectural aspects of the proposed Deep Network, specifically designed for saliency analysis in driving scenarios.

To capture the driving scenario, we utilized an automotive-grade camera device, which enabled the acquisition of video frames. Subsequently, we employed a custom-designed 3D-to-2D Semantic Segmentation Fully Convolutional Network (SS-FCN) to process the collected video frames and extract saliency information. The architecture of our proposed segmentation model follows an encoder/decoder structure, facilitating efficient processing of the driving scene. The resulting saliency map highlights the most prominent object within the captured driving scenario. Our SS-FCN architecture consists of two primary components: an encoder and a decoder. The encoder block, known as 3D Enc Net, focuses on extracting spatiotemporal features. It comprises five blocks, with each block consisting of convolutional layers grouped into two sets of two layers. The kernel size for each separable convolutional layer is 3x3x3. Within each block, a sequence of two 3x3x3 convolutional operations is followed by batch normalization, a Rectified Linear Unit (ReLU) layer, and a max pooling operation with a pooling size of 1x2x2. This sequence is repeated twice. Subsequently, another sequence of 3x3x3 convolutional layers is performed. The remaining three blocks follow a similar pattern, with a succession of two convolutional operations, followed by batch normalization, another convolutional layer with a 3x3x3 kernel, batch normalization, and ReLU activation, culminating in a downstream max-pooling layer with a size of 1x2x2.
The decoder backbone, referred to as 2D Dec Net, comprises five blocks and incorporates up-sampling layers for feature map decoding. It consists of 2D convolutional layers with a kernel size of 3x3. After each convolutional layer, batch normalization and ReLU activation are applied. Residual connections are introduced through convolutional blocks. The decoder section is responsible for adjusting the size of the feature maps through up-sampling operations. Ultimately, the output of our proposed SS-FCN architecture is the feature map corresponding to the segmented area of the most salient object within the driving scenario's video frame. 
Semantic segmentation techniques often highlight the most visually salient objects in an image, typically represented by fixation points. To illustrate this, we present the saliency maps of the driving scenarios and the corresponding training loss dynamics of the SS-FCN model in Figures 3 and 2, respectively. For training and testing the SS-FCN model, we utilized the DHF1K dataset [1]. Performance evaluation of the model was conducted using several performance indexes, including the Area Under the Curve (AUC), Similarity, Correlation Coefficient, and Normalized Scanpath Saliency. The proposed architecture demonstrated satisfactory performance, with the following results: AUC: 0.885, Similarity: 0.355, Correlation Coefficient: 0.455, and Normalized Scanpath Saliency: 2.564. These values compare favorably with similar architectures found in the literature [1]. It is worth noting that while other architectures may exhibit superior performance, they often require high computational resources and rely on sophisticated hardware. In contrast, our proposed pipeline achieves notable results while maintaining a low computational workload and without the need for specific hardware accelerations [2].
\begin{figure}
\centerline{\includegraphics[ width=0.9\columnwidth]{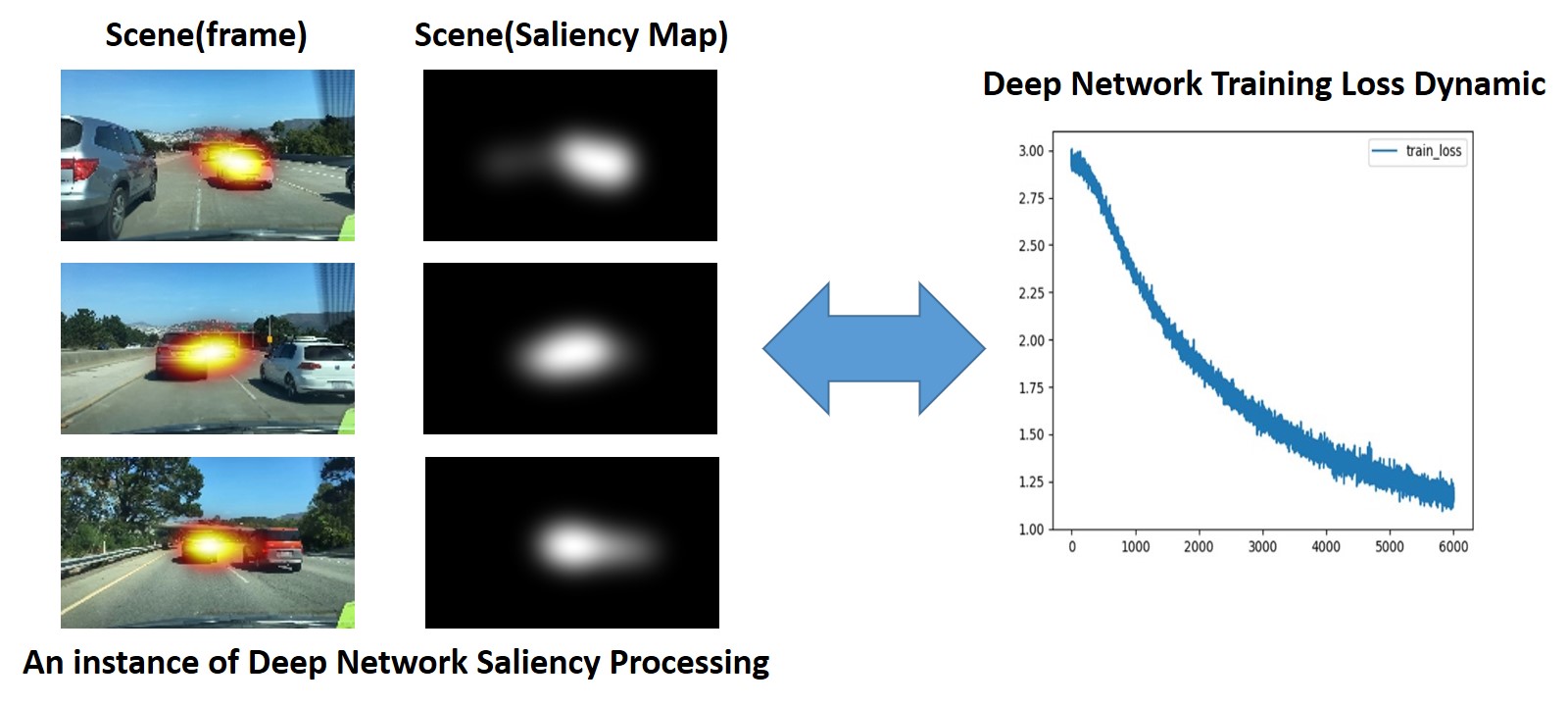}}
\caption{An instance of saliency processing made by fully trained SS-FCN. The network training loss dynamic is reported}
\label{fig2}
\end{figure}

\begin{figure*}
\centerline{\includegraphics[ width=0.8\textwidth]{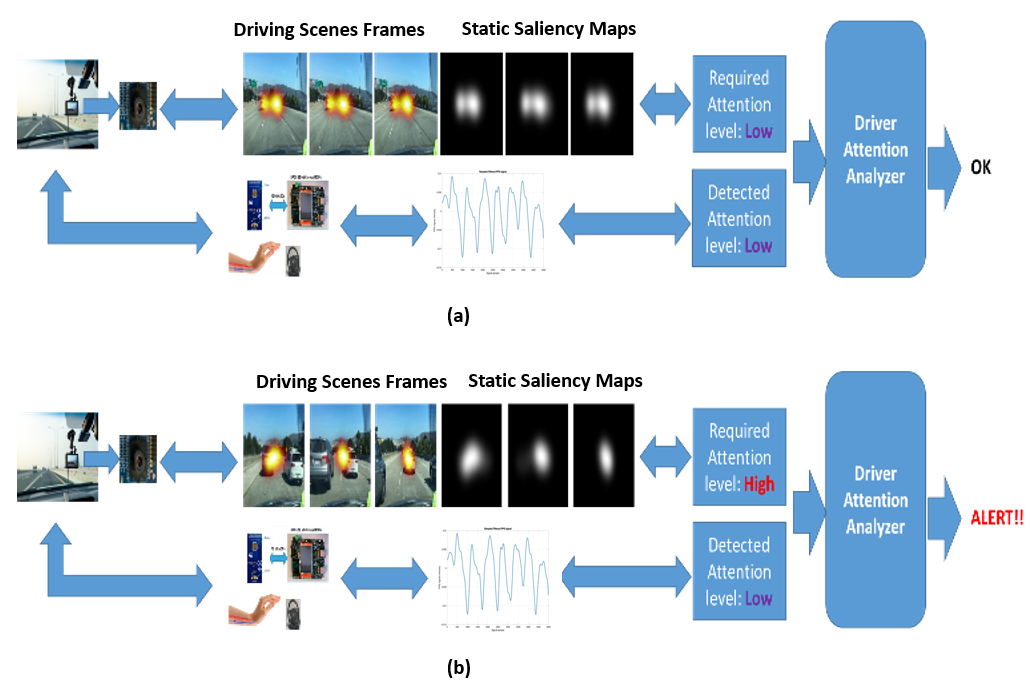}}
\caption{(a) Static driving scenario requiring low attention level. The Driver Attention Analyzer confirms an adequate level of attention (b) Dynamic driving scenario (car overtaking) requiring High attention level. The Driver Attention Analyzer detects an inadequate level of attention generating an acoustic signal alert.}
\label{fig4}
\end{figure*}

\subsection{The Driver Attention Analyzer}
This component serves to evaluate the driver's attention level by combining the analysis of the PPG signal with the output of the proposed SS-FCN block, namely the saliency map. Its main objective is to ensure that the detected attention level aligns with the meaningful interpretation of the corresponding salience map, which represents the dynamic nature of the driving scenario. A static saliency map indicates a relatively low level of dynamic activity in the driving scenario, suggesting a corresponding lower level of attention from the driver. On the other hand, an increasing variation in the saliency maps enable a greater demand for the driver's attention. Hence, this block verifies the coherence between the driver's attention level and the saliency map, reflecting the dynamic characteristics of the driving scenario \cite{cai2017saliency}. We set ad-hoc thresholds to verify the level of attention obtained by the Drowsiness Monitoring systems. Both thresholds are determined during the training phase by means of a heuristic calibration i.e. by choosing a setup that maximizes the performance of the overall pipeline during the learning.

\section{Experimental Results}

In this section, we begin by introducing the dataset and then proceed to discuss the training details and the key findings of our study. To evaluate the effectiveness of our implemented pipeline, we utilized the DH1FK dataset [1]. Additionally, we captured video sequences of various driving scenarios using a camera with a resolution of 2.3 megapixels and a maximum frame rate of 60 frames per second. The recruitment of subjects was supervised by a group of physiologists who ensured that the recruited individuals exhibited diverse levels of attention while simultaneously collecting the corresponding EEG signal, thereby corroborating their physiological responses [2].
Our experiments involved a total of 43 subjects ranging in age from 21 to 70 years. We recorded the PPG signal from each subject at a sampling frequency of 1 KHz, with a data collection duration of 5 minutes. For training purposes, we allocated 70\% of the acquired PPG time-series and video driving scene frames, reserving the remaining 30\% for testing and validation.  In summary, our proposed pipeline aims to assess the driver's attention level by combining the PPG signal and the saliency map. Specifically, the Driver Attention Analyzer block compares the level of attention required by the driving scenario with the output of the Drowsiness Monitoring System. If the driver's vigilance level falls below the requirement of the current driving scenario, the system alerts the driver through an acoustic signal. To classify the attention level, we defined two value ranges. A value within the range of 0 to 0.6, which is the output of the 1D-CNN-based Drowsiness Monitoring System, indicates a medium-low attention level. Conversely, values ranging from 0.61 to 1 represent a high attention level.  Finally, we set an ad-hoc normalized threshold $\vartheta$ (0.45) to define a static scene-based saliency map (see Eqs. (1)). The driving scene is considered as ‘dynamic’ if the values of the normalized saliency map gradient are greater than $0.45$ (requiring a high level of attention). For a normalized saliency map gradient lower than $0.45$ (requiring a low level of attention) the driving scene is considered ‘static’. In Fig.\ref{fig4}, two instances of the driving scenarios are reported.

\section{Conclusions and Discussion}
The promising results obtained from our study highlight the potential for assessing driver drowsiness and enhancing driving safety. A key advantage of our proposed method is its independence from frequency domain analysis, unlike other approaches that rely on HRV analysis [1]. Additionally, our method leverages the readily available PPG signal, which can be conveniently sampled from a range of sensors placed on the steering wheel.
To evaluate the driver's level of attention based on PPG signal analysis, we developed a fully convolutional deep network that generates a saliency map of the driving scene. The results demonstrate the effectiveness of our pipeline in assessing the required level of driver drowsiness specific to the driving scenario. The system compares the attention level reconstructed from the driver's PPG signal with the calibrated attention level derived from saliency analysis. If a discrepancy is detected between these attention levels, indicating a potential risk, the system alerts the driver. Ongoing investigations aim to enhance the saliency analysis by expanding the dataset to include diverse driving scenarios from various domains. Currently, we are in the process of porting the entire pipeline to an automotive-specific board powered by the SoC STA1295 ACCORDO 5 processor manufactured by STMicroelectronics [2]. Furthermore, our future endeavors involve incorporating more robust domain adaptation methods, encompassing both supervised and unsupervised approaches that have been successfully employed in other pipelines [3][4][5][6].

\section*{Acknowledgment}
The used dataset was collected under the clinical study Ethical Committee CT1 authorization n.113 / 2018 / PO. This research was funded by the National Funded Program 2014-2020 under grant agreement n. 1733, (ADAS + Project).

\bibliographystyle{IEEEtran.bst}
\bibliography{refs}

\end{document}